\documentclass[12pt]{article}

\usepackage{graphicx}
\usepackage{amssymb}
\usepackage{amsmath}
\usepackage{bbm}

\newcommand{\real}{\mathbbm{R}}

\begin{document}
\begin{center}
{\bf \Large Machine Learning for Initial Value Problems \\[0.5ex] 
of Parameter-Dependent Dynamical Systems}

\vspace{10mm}

{\large Roland Pulch${}^1$ and Maha Youssef${}^2$}

\bigskip

{${}^1$Universit\"at Greifs\-wald,
  Institute of Mathematics and Computer Science,
  Walther-Rathenau-Str.~47, D--17489 Greifs\-wald, Germany. \\
  Email: {\tt roland.pulch@uni-greifswald.de}}  

\bigskip

{${}^2$Universit\"at Stuttgart,
  Institute of Applied Analysis and Numerical Simulation,
  Pfaffenwaldring~57, D--70569 Stuttgart, Germany. \\
  Email: {\tt maha.youssef-ismail@mathematik.uni-stuttgart.de}} 
\end{center}

\bigskip

\begin{center}
{\bf Abstract}

\bigskip

\begin{tabular}{p{12.5cm}}
  We consider initial value problems of nonlinear dynamical systems,
  which include physical parameters.
  A quantity of interest depending on the solution is observed.
  A discretisation yields the trajectories of the quantity of interest
  in many time points.
  We examine the mapping from the set of parameters to the discrete values
  of the trajectories.
  An evaluation of this mapping requires to solve an initial value problem.
  Alternatively, we determine an approximation, where the evaluation
  requires low computation work, using a concept of machine learning.
  We employ feedforward neural networks,
  which are fitted to data from samples of the trajectories.
  Results of numerical computations are presented for a test example
  modelling an electric circuit.
\end{tabular}
\end{center}


\section{Introduction}
\label{pulch:sec:intro}
We examine initial value problems of nonlinear dynamical systems
consisting of ordinary differential equations (ODEs)
or differential-algebraic equations (DAEs).
The systems include physical parameters, which vary in a predetermined 
bounded domain.
A quantity of interest (QoI) is defined depending on the solution of
the dynamical system.
Hence there is a mapping from the parameters onto the trajectories
of the QoI in the time domain.
Our aim consists in the determination of an approximation of
this mapping, which can be evaluated with a low computational effort.
These approximations can be applied as surrogate models in
uncertainty quantification, see~\cite{pulch:lemaitre-knio}, for example.

Methods using polynomials and their orthogonal bases yield surrogate models
of parametric problems, see~\cite{pulch:pulch-scee2008,pulch:sullivan}.
Alternatively, we employ an approach of machine learning using artificial
neural networks (NNs), see~\cite{pulch:du-swamy,pulch:goodfellow-book},
to construct an approximation.
In~\cite{pulch:wang-etal,pulch:yu-hesthaven},
proper orthogonal decomposition (POD) is applied to data of solutions
of parametric partial differential equations to obtain a reduced basis.
Consequently, a mapping between low-dimensional spaces is approximated
by NNs.
In contrast, we discretise in time and use the data from the trajectories
of the QoI in many time points.
A mapping from a low-dimensional parameter space to a high-dimensional
space is approximated by an NN now.

We apply feedforward NNs for this approximation.
The determination of an NN requires a training procedure using data.
We obtain the data of the trajectories by solving initial value problems
of the dynamical systems for samples of the parameters.
The fitting of an NN represents a (nonlinear) optimisation problem.
In an NN, the number of neurons in a hidden layer is typically larger than
the number of neurons in the input layer or the output layer.
Since the number of outputs is large in our case,
we also have high numbers of neurons in the hidden layers.

Finally, we demonstrate numerical results for a test example,
which is a DAE model of an electric circuit.
The trajectories associated to some parameter samples are illustrated.
We show statistics of the approximation errors.


\section{Parameter-dependent Dynamical Systems}
\label{pulch:sec:problem}
Let parameters ${p} \in \Pi \subset \real^q$ be given.
We consider a nonlinear dynamical system of the form
\begin{equation}
  \label{pulch:eq:dae}
  {M}({p}) \dot{{x}}(t,{p}) = 
    {f}(t,{x}(t,{p}),{p}) .
\end{equation}
The mass matrix ${M} : \Pi \rightarrow \real^{n \times n}$ and
the right-hand side
${f}: [t_0,t_{\rm f}] \times \real^n \times \Pi \rightarrow \real^n$
include the parameters.
Thus the solution ${x} : [t_0,t_{\rm f}] \times \Pi \rightarrow \real^n$
depends both on time and the parameters.
If the mass matrix is non-singular, then~(\ref{pulch:eq:dae})
represents a system of ODEs.
If the mass matrix is singular, then a system of DAEs is given.
We examine initial value problems (IVPs)
\begin{equation} \label{pulch:eq:initial-values}
  {x}(t_0,{p}) = {x}_0({p}) .
\end{equation}
In the case of DAEs, the initial values have to be consistent,
see~\cite{hairer-wanner-2}.
Consistent initial values often depend on the parameters. 
We define a QoI
$y : [t_0,t_{\rm f}] \times \Pi \rightarrow \real^n$ by a function
$g : \real^n \rightarrow \real$ via
\begin{equation} \label{pulch:eq:qoi}
  y(t,{p}) = g({x}(t,{p})) .  
\end{equation}
Each selection of the parameters yields a trajectory of the QoI
in the time domain.
We obtain the mapping
\begin{equation} \label{pulch:eq:trajectories}
  {p} \mapsto \{ (t,y(t,{p})) \; : \; t \in [t_0,t_{\rm f}] \}
\end{equation}
for any ${p} \in \Pi$.
Our aim is to construct an approximation of the
mapping~(\ref{pulch:eq:trajectories}), which can be evaluated cheap,
in particular, 
without solving IVPs (\ref{pulch:eq:dae}),(\ref{pulch:eq:initial-values})
any more.

The following approach can also be used for boundary value problems (BVPs)
of dynamical systems, because we only include the trajectories of the QoI
in the method.
The trajectories are computed from either IVPs or BVPs. 


\section{Time Discretisation}
\label{pulch:sec:time}
We discretise the trajectories of the QoI~(\ref{pulch:eq:qoi}) in
the time domain $[t_0,t_{\rm f}]$.
Let
\begin{equation} \label{pulch:eq:time-points}
  t_0 < t_1 < t_2 < \cdots < t_{m-1} < t_m \le t_{\rm f} .
\end{equation}
Equidistant time points can be used. 
We consider the mapping $\Theta : \Pi \rightarrow \real^m$
\begin{equation} \label{pulch:eq:map-disc}
  {\Theta} : \Pi \rightarrow \real^m , \quad
  \Theta({p}) =
  \left( \begin{array}{c}
    y(t_1,{p}) \\
    \vdots \\
    y(t_m,{p}) \\
  \end{array} \right) .
\end{equation}
Each evaluation of~(\ref{pulch:eq:map-disc}) requires to solve
an IVP (\ref{pulch:eq:dae}),(\ref{pulch:eq:initial-values})
followed by the extraction of the QoI~(\ref{pulch:eq:qoi}).
The IVPs of the dynamical systems are solved by numerical methods,
see~\cite{hairer-wanner-1,hairer-wanner-2},
like Runge-Kutta schemes and linear multistep methods.
The methods yield approximations of the solution in discrete time points,
which are typically determined by a local error control.
Thus these time points are not identical to our
choice~(\ref{pulch:eq:time-points}).
Nevertheless, we obtain the solution in the
points~(\ref{pulch:eq:time-points}) by an interpolation
or a dense output in time.

Stiff systems of ODEs and all DAEs require implicit methods in
the time integration.
Therein, a nonlinear system of algebraic equations has to be solved
in each time step.
Thus the computational effort becomes large.
Our goal is to determine an approximation of the
mapping~(\ref{pulch:eq:map-disc}), whose evaluation is cheap.


\section{Machine Learning}
\label{pulch:sec:nn}
\begin{figure}[t!]
\centering
\includegraphics[width=8cm]{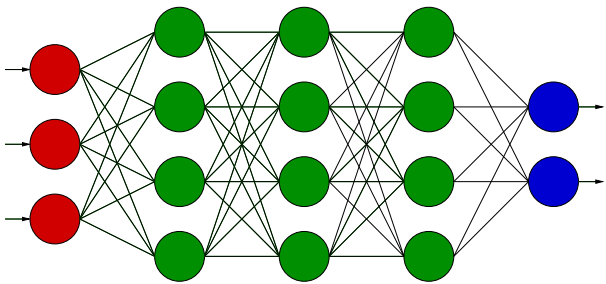}
\caption{Artificial NN with input layer (red),
  hidden layers (green), and output layer (blue).}
\label{pulch:fig:neuralnet}       
\end{figure}

We arrange an artificial NN, see~\cite{pulch:goodfellow-book},
to approximate the mapping~(\ref{pulch:eq:map-disc}). 
An NN consists of an input layer, an output layer and
additional hidden layers.
Figure~\ref{pulch:fig:neuralnet} illustrates the schematic of an NN.
Let $N_j$ be the number of neurons in the $j$th layer for
$j=0,1,\ldots,J$.
Therein, $j=0$ and $j=J$ represent the input layer and the
output layer, respectively.
Thus there are $J-1$ hidden layers.
It holds that $N_0=q$ and $N_{J}=m$ and thus $N_J \gg N_0$ in our problem.

The mathematical model $\Psi : \real^{N_0} \rightarrow \real^{N_{J}}$
of an NN consists in a chain of operators 
\begin{equation} \label{eq:neural-network}
  \Psi =
  {T}_{J} \circ \rho \circ {T}_{J-1} \circ \rho \circ {T}_{J-2}
  \circ \cdots \circ \rho \circ {T}_2 \circ \rho \circ {T}_1 .
\end{equation}
Each operator ${T}_j : \real^{N_{j-1}} \rightarrow \real^{N_{j}}$
is an affine-linear function
$$ {T}_{j} ({z}) = {A}_{j} {z} + {b}_{j} $$
including a matrix ${A}_{j} \in \real^{N_{j} \times N_{j-1}}$,
a vector ${b}_{j} \in \real^{N_{j}}$,
and the input~${z} \in \real^{N_{j-1}}$.
In the context of machine learning,
the entries of~${A}_{j}$ and ${b}_{j}$ are denominated as
weights and biases, respectively.
The operator~$\rho$ is a nonlinear transfer function
$\rho : \real \rightarrow \real$
(also called activation function).
Typical choices are, for example,
the hyperbolic tangent sigmoid function
\begin{equation} \label{pulch:eq:tansig}
  \rho (x) = \frac{2}{1+{\rm e}^{-2x}} - 1
\end{equation}
and the hard-limit function
\begin{equation} \label{pulch:eq:hardlim}
  \rho (x) = \left\{ \begin{array}{ll}
  0 & \;\;\mbox{for}\; x < 0 , \\
  1 & \;\;\mbox{for}\; x \ge 0 . \\
  \end{array} \right.
\end{equation}
In~(\ref{eq:neural-network}), the function~$\rho$ is applied to
vectors in each component separately.

In the fitting of an NN, the ideal is to minimise the distances
$\Theta({p}) - \Psi({p})$ for any~${p} \in \Pi$.
The degrees of freedom are the weights and biases, i.e.,
$({A}_j,{b}_j)_{j=1}^J$.
Since a nonlinear optimisation problem appears,
iterative methods are required to obtain numerical solutions.

In practise, the fitting involves three sample sets for training,
validation, and test:
\begin{equation} \label{pulch:eq:sample-sets}
  \begin{array}{rl}
  \mathcal{S}_{\rm train} & = \{ {p}_1, \ldots ,{p}_{k} \}
  \subset \Pi  \\[0.5ex] 
  \mathcal{S}_{\rm valid} & = \{ {q}_1, \ldots ,{q}_{k'} \}
  \subset \Pi  \\[0.5ex] 
  \mathcal{S}_{\rm test} & = \{ {r}_1, \ldots ,{r}_{k''} \}
  \subset \Pi . \\ 
  \end{array}
\end{equation}
For example, random samples can be chosen,
where a uniform probability distribution is assumed in the
parameter domain~$\Pi$.
The minimisation is based on the differences
$\Theta({p}_i) - \Psi({p}_i)$ for parameter tuples~${p}_i$
from the training set.
The (vector-valued) differences are measured using the mean squared error
or the mean absolute error.
The error measure decreases monotone for the parameters in the training set
due to the minimisation.
The validation set is included to prevent an overfitting.
If the error measure of the validation set increases, then the training
is stopped and the best previous case is put out.
The test set is not involved in the minimisation at all.
Hence this set allows for an estimate of the quality of the trained~NN.


\section{Numerical Results for Test Example}
\label{pulch:sec:example}
All numerical computations were performed within the software
MATLAB~\cite{pulch:matlab2019} using the Deep Learning Toolbox.

\begin{figure}[t]
  \centering
  \includegraphics[width=6cm]{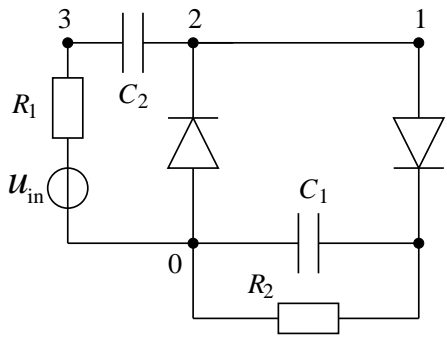}
  \caption{Diagram of electric circuit.}
  \label{pulch:fig:circuit}       
\end{figure}

We investigate an electric circuit introduced in~\cite{pulch:barz},
which is illustrated by Fig.~\ref{pulch:fig:circuit}.
This circuit performs a voltage doubling for specific choices of
parameters and input voltage.
A mathematical modelling yields a nonlinear system of DAEs~(\ref{pulch:eq:dae})
with $n=3$ equations for the three unknown node voltages
presented in~\cite{pulch:barz}:
\begin{equation} \label{pulch:eq:dae-example}
  \begin{array}{rcl}
    C_1 \dot{x}_1 & = & - \frac{x_1}{R_2} + F(-(x_1+x_3)) \\
    C_2 \dot{x}_2 & = & - \frac{1}{R_1} (x_2+x_3+u_{\rm in}) \\
    0 & = & - \frac{1}{R_1} (x_2+x_3+u_{\rm in}) + F(-(x_1+x_3)) - F(x_3) . \\
  \end{array}
\end{equation}
The current-voltage relation of the diodes reads as
$F(u) = \gamma (\exp(\delta u)-1)$.
We use the constant parameters $\gamma = 4.067 \cdot 10^{-8}$
and $\delta = 5.634 \cdot 10^{-2}$ given in~\cite{pulch:kampowsky-etal}.
The differential index of the DAE system~(\ref{pulch:eq:dae-example}) is one.
We choose the second node voltage as QoI~(\ref{pulch:eq:qoi}).

We consider variations in four physical parameters:
the two capacitances and the two resistances. 
The ranges $C_j \in [2\cdot 10^{-9},3\cdot 10^{-9}]$
for $j=1,2$, $R_1 \in [10^6,2\cdot 10^6]$, $R_2 \in [10^8,2\cdot 10^8]$
form the parameter domain~$\Pi \subset \real^4$.

\begin{figure}
  \centering
  \includegraphics[width=6cm]{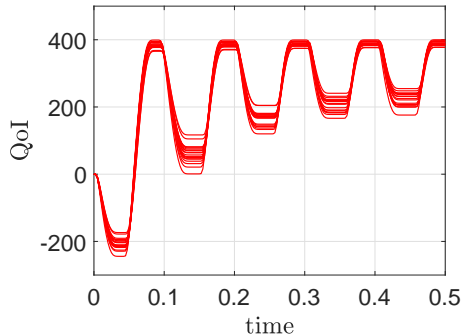}
  \caption{Twenty trajectories of the QoI for different parameter samples.}
  \label{pulch:fig:variation}       
\end{figure}

As input voltage, we supply the harmonic oscillation
$$ u_{\rm in}(t) = A \sin \left( \frac{2\pi}{T} \, t \right) $$
with amplitude $A=500$ and period $T=0.1$.
The total time interval of our simulations is $[t_0,t_{\rm f}]=[0,0.5]$.
The initial values~(\ref{pulch:eq:initial-values}) are set to zero, 
which represents a consistent case in this example.
The backward differentiation formulas (BDF), see~\cite{hairer-wanner-1},
yield the numerical solutions of the IVPs.
High accuracy requests are imposed in the local error control
with relative tolerance $\varepsilon_{\rm rel}=10^{-4}$
and absolute tolerance $\varepsilon_{\rm abs}=10^{-6}$. 
The error control generates approximations on a non-uniform grid in time.
We extract the trajectories of the QoI in $m=200$ equidistant
time points $t_{\ell}=\ell \Delta t$ for $\ell = 1,\ldots,m$
with $\Delta t = \frac{t_{\rm f}-t_0}{m}$ by interpolation.
The order of accuracy coincides for both the uniform grid and non-uniform grid.
The associated error of the time integration is negligible in comparison 
to the approximation error of the NNs below.
Figure~\ref{pulch:fig:variation} gives an impression of the
variability within the trajectories of the QoI for our 
parameter domain.

\begin{figure}
\centering
\includegraphics[width=6.2cm]{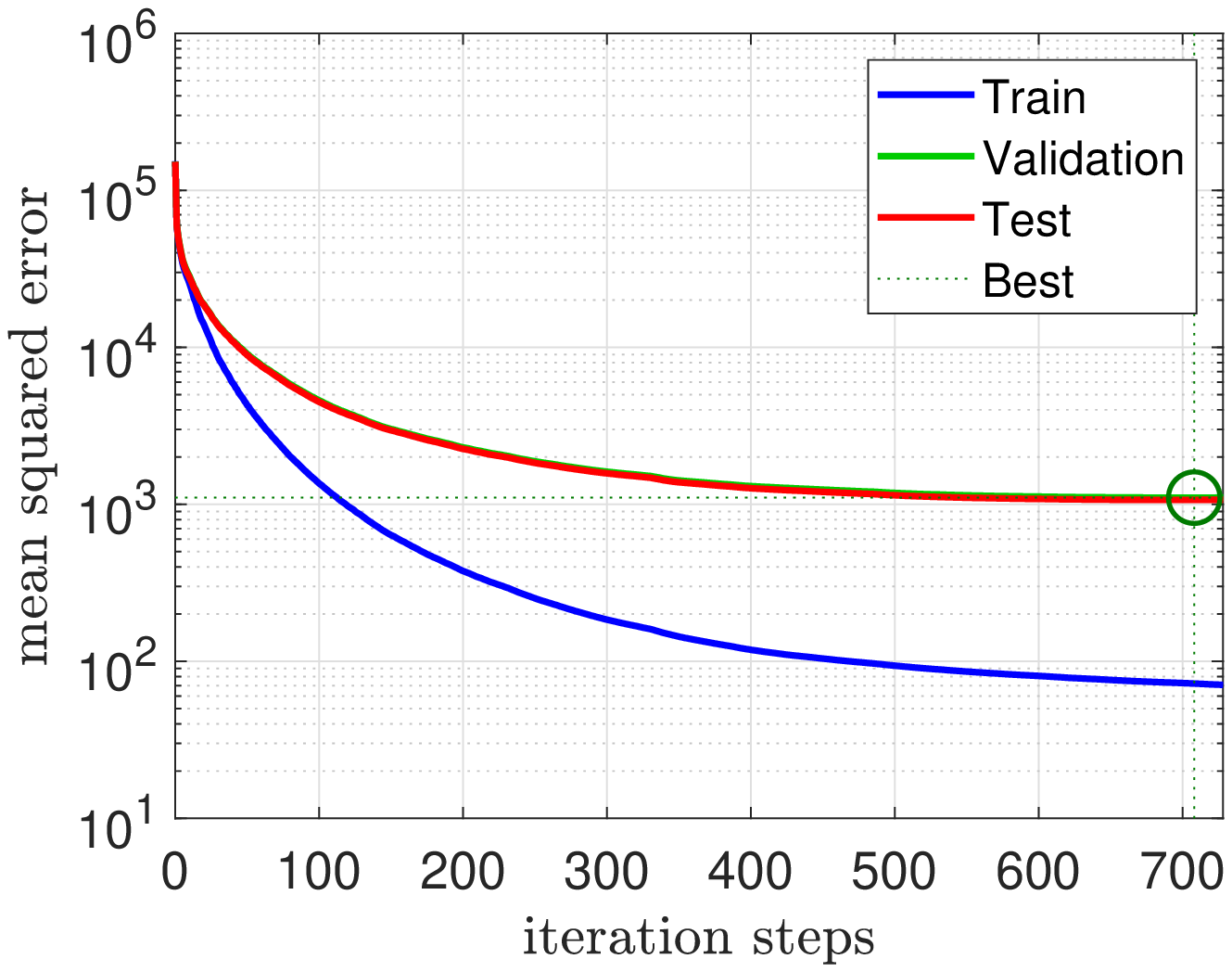}
\hspace{5mm}
\includegraphics[width=6.2cm]{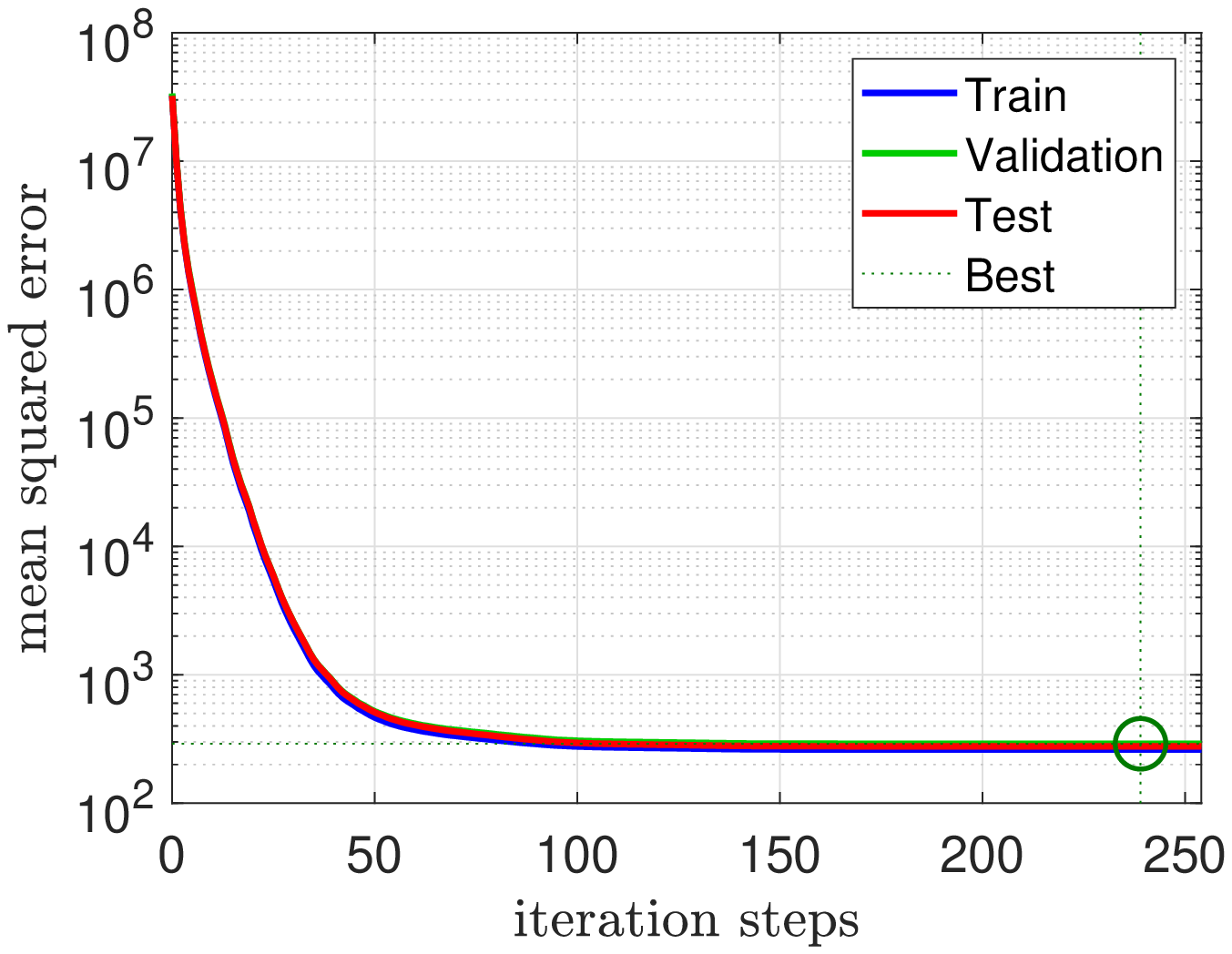}

(i) \hspace{6cm} (ii)
\caption{Mean squared errors during the fitting of the two NNs
  in the iterative minimisation. 
  (The green line of the validation set is mostly located behind
  the red line of the test set.)}
\label{pulch:fig:performance}       
\end{figure}

We select the number of samples as $k=k'=k''=500$
in the sets~(\ref{pulch:eq:sample-sets}). 
Often the validation set and the test set are chosen smaller than 
the training set due to a restricted amount of data. 
In contrast, we are able to use larger sets, since a high number of 
trajectories can be produced by numerical simulations. 
In particular, a large test set provides reliable statistics 
in the error analysis.
A pseudo random number generator yields the parameter samples
in the multi\-dimen\-sional cuboid~$\Pi$. 
Our NNs include two hidden layers with 400~neurons in each layer. 
Using more hidden layers or more neurons did not improve the 
results significantly.
We investigate two NNs, which differ only in the choice of the
transfer function:
\begin{itemize}
\item[i)]
  hard-limit transfer function~(\ref{pulch:eq:hardlim}),
\item[ii)]
  purely linear transfer function.
\end{itemize}
In the training, a conjugate gradient backpropagation method
iteratively solves the minimisation problem. 
Figure~\ref{pulch:fig:performance} shows the performance of the
training procedure.
In the case of the hard-limit transfer function, the training is stopped
at the 728th iteration step, because the error of the validation set
increases slightly.
In the case of the linear transfer function, the training is terminated
at the 254th iteration step due to a too small step size. 
These two NNs are used in the following error analysis.

In addition, we tried two other backpropagation techniques in the training 
of the NNs: 
a one-step secant method (quasi Newton method) and 
a gradient descent method with momentum and adaptive learning rate.
More information on all three methods can be found in~\cite{pulch:du-swamy}, 
for example.
Table~\ref{pulch:tab:training} demonstrates the number of iteration steps 
(until a termination criterion applies) as well as the final mean squared error 
of the test set for the three techniques. 
We observe that the conjugate gradient method exhibits the best performance. 

\begin{table}[h]
  \caption{Number of iteration steps and mean squared error (MSE) of test set 
  for different training methods in NNs with hard-limit transfer function~(i) 
  and purely linear transfer function~(ii).}
  \label{pulch:tab:training}
  \centering

  \medskip

  \begin{tabular}{lcccc} \hline
  & \multicolumn{2}{c}{(i)} & \multicolumn{2}{c}{(ii)} \\
    & steps & \quad MSE \quad \mbox{} & steps & \quad MSE \quad \mbox{} \\ \hline
    conjugate gradient method \quad \mbox{} & 728 & 1067.5 & 254 & 277.21 \\
    one-step secant method & 1696 & 1062.2 & 328 & 277.24 \\
    gradient descent method & 10000 & 1323.1 & 1413 & 277.74 \\ \hline
  \end{tabular}
\end{table}

\begin{figure}
  \centering
  \includegraphics[scale=.99]{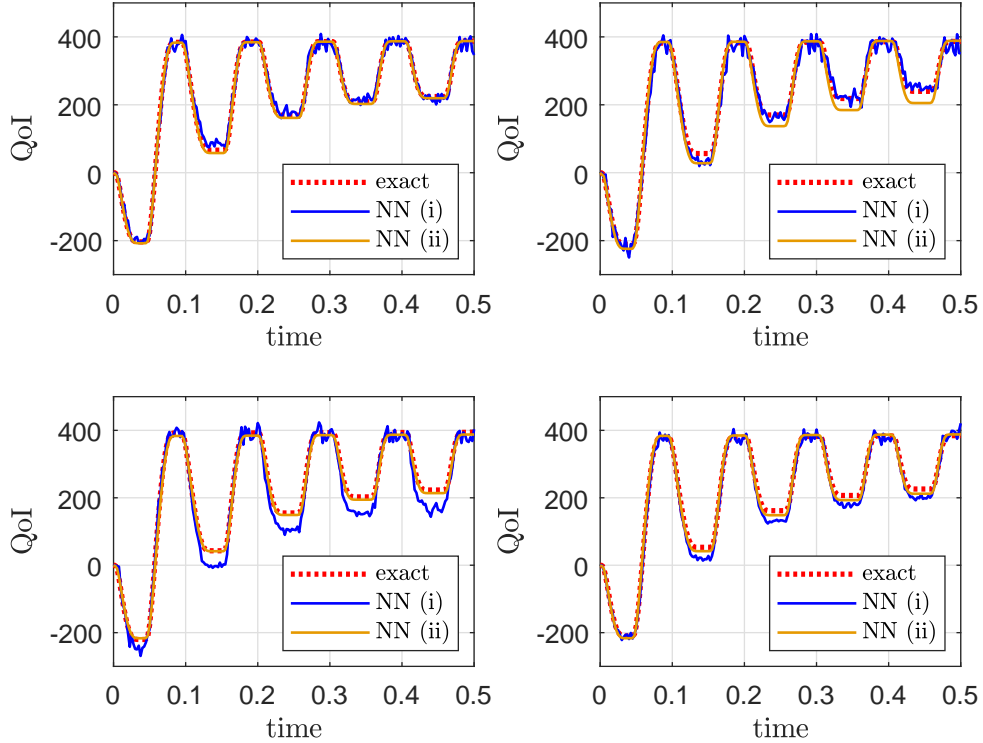}
  \caption{Trajectories of the QoI for some samples from the test set.}
  \label{pulch:fig:samples}       
\end{figure}

Figure~\ref{pulch:fig:samples} illustrates several trajectories of the
test set.
A comparison of the exact trajectories from the time integration
and the approximations from the two NNs is shown.
An interesting property is that NN~(i) with the nonlinear transfer function
exhibits an oscillatory behaviour of the approximations in time,
whereas NN~(ii) with the linear transfer function generates more smooth
approximations in time.
The incorrect behaviour is not a transient effect,
because all discrete time points~(\ref{pulch:eq:time-points})
are treated equally in the NNs, i.e.,
without the consideration of their ordering in time.
We also tried other nonlinear transfer functions
(like~(\ref{pulch:eq:tansig})),
which still caused oscillations in time.

Finally, we quantify the relative errors of the approximations
for each sample trajectory using a discrete $\mathcal{L}^1$-norm
in time.
The error associated to the $i$th parameter sample reads as
\begin{equation} \label{pulch:eq:error}
  E_i = \frac{t_{\rm f}-t_0}{m} \sum_{\ell=1}^m
  \frac{\left| \tilde{y}(t_{\ell},{p}_i) - y(t_{\ell},{p}_i) \right|}{
    | y(t_{\ell},{p}_i)|},
\end{equation}
where $y$ is the original value from the time integration
and~$\tilde{y}$ denotes the approximation from an NN.
The initial value is not included due to its value zero.
The statistics of the errors are depicted in Table~\ref{pulch:tab:error}.
We discuss the resulting mean values.
In NN~(i), a smaller mean error is achieved in the training set,
whereas the other two sets show larger errors in comparison to NN~(ii).
Moreover, the mean errors are balanced for all three sets in NN~(ii).
This behaviour is in agreement to the performance of the
training demonstrated by Fig.~\ref{pulch:fig:performance}. 

\begin{table}
  \caption{Mean value and standard deviation of relative errors in
    discrete $\mathcal{L}^1$-norm, see~(\ref{pulch:eq:error}),
    for the three parameter sets within the two trained NNs.}
  \label{pulch:tab:error}
  \centering

  \medskip

  \begin{tabular}{clccc} \hline
    & & NN (i) & \mbox{}\quad\mbox{} & NN (ii) \\ \hline
    mean \quad \mbox{} & training set & 0.044 & & 0.086 \\
    & validation set\quad \mbox{} & 0.130 & & 0.086 \\
    & test set & 0.120 & & 0.079 \\ \hline
  \end{tabular}
  \hspace{5mm}
  \begin{tabular}{clccc} \hline
    & & NN (i) & \mbox{}\quad\mbox{} & NN (ii) \\ \hline
    st.dev.\quad\mbox{} & training set & 0.135 & & 0.228 \\
    & validation set\quad \mbox{} & 0.210 & & 0.158 \\
    & test set & 0.155 & & 0.146 \\ \hline
  \end{tabular}
\end{table}


\section{Conclusions}
We arranged a mapping from a set of parameters to discrete values of
a QoI obtained from IVPs of dynamical systems.
We approximated this mapping by artificial NNs.
A test example was investigated, where two NNs were trained.
In both NNs, the quality of the approximations is moderate with respect
to the mean values of the errors.
However, the NN including a linear transfer function yielded more smooth
discretised trajectories in time,
whereas the NNs with nonlinear transfer functions produced
incorrect oscillations in time.

\clearpage




\end{document}